\pdfoutput=1

\documentclass[11pt]{article}

\usepackage{tikz}
\usetikzlibrary{positioning, patterns}

\usepackage{pgfplots}
\usepackage{pgfplotstable}
\usepackage{pgfplots}
\pgfplotsset{compat=1.18}
\usetikzlibrary{patterns}

\usepackage{times}
\usepackage{latexsym}

\usepackage[T1]{fontenc}

\usepackage[utf8]{inputenc}

\usepackage{microtype}

\usepackage{inconsolata}

\usepackage{graphicx}

\usepackage{enumitem}
\usepackage{booktabs}
\usepackage{amsmath}

\setlist[itemize]{noitemsep, topsep=0pt, parsep=0pt, partopsep=0pt}

\usepackage[most]{tcolorbox}
\tcbuselibrary{listings,breakable}

\newtcblisting{promptbox}{
  breakable,
  listing only,
  width=\columnwidth,
  colback=gray!5,
  colframe=black!60,
  boxrule=0.5pt,
  arc=2pt,outer arc=2pt,
  left=4pt,right=4pt,top=4pt,bottom=4pt,
  before skip=8pt,after skip=8pt,
  listing options={
    basicstyle=\ttfamily\small,
    breaklines=true,
    breakautoindent=false,  
    breakindent=0pt,       
    columns=fullflexible,
    keepspaces=true,
    xleftmargin=0pt,       
    framexleftmargin=0pt,   
    gobble=0                
  },
}

\usepackage{tabularx,array}
\newcolumntype{L}{>{\raggedright\arraybackslash}X}

\newtcolorbox{instrbox}{
  breakable, enhanced, colback=gray!2, colframe=black!40,
  boxrule=0.5pt, arc=2pt, outer arc=2pt,
  left=6pt,right=6pt,top=6pt,bottom=6pt, title=\textbf{Annotator Instructions}
}

\newtcolorbox{transcriptbox}{
  breakable, enhanced, colback=gray!3, colframe=black!25,
  boxrule=0.5pt, arc=2pt, outer arc=2pt,
  left=6pt,right=6pt,top=6pt,bottom=6pt
}

\newtcolorbox{infcard}[1][]{%
  breakable, enhanced,
  colback=white, colframe=black!20,
  boxrule=0.5pt, arc=2pt, outer arc=2pt,
  borderline west={2pt}{0pt}{black!60},
colbacktitle=white, coltitle=black,
  titlerule=0pt,
  left=6pt,right=6pt,top=6pt,bottom=6pt,
  #1
}

\newcommand{\speaker}[1]{\textbf{#1:}}

\newtcolorbox{scenebox}{
  breakable, enhanced, colback=gray!1, colframe=black!40,
  boxrule=0.5pt, arc=2pt, outer arc=2pt,
  left=6pt,right=6pt,top=6pt,bottom=6pt
}
\newtcolorbox{llmbox}[1][]{breakable, enhanced,
  colback=white, colframe=black!25, boxrule=0.5pt,
  borderline west={2pt}{0pt}{black!40},
  arc=2pt, outer arc=2pt, title=\textbf{#1}
}
\newtcolorbox{lrmbox}[1][]{breakable, enhanced,
  colback=white, colframe=black!25, boxrule=0.5pt,
  borderline west={2pt}{0pt}{black!70},
  arc=2pt, outer arc=2pt, title=\textbf{#1}
}
\newtcolorbox{subsupport}{
  breakable, enhanced, colback=gray!1, colframe=black!20,
  boxrule=0.4pt, arc=2pt, outer arc=2pt, title=\scriptsize\textbf{Supporting}
}
\newtcolorbox{suboppose}{
  breakable, enhanced, colback=gray!1, colframe=black!20,
  boxrule=0.4pt, arc=2pt, outer arc=2pt, title=\scriptsize\textbf{Opposing}
}

\newenvironment{infcheck}{
  \begin{itemize}[leftmargin=*, itemsep=2pt, topsep=4pt, parsep=0pt]
}{\end{itemize}}

\usepackage[preprint]{acl}

\title{SocialNLI: A Dialogue‐Centric Social Inference Dataset}

\author{Akhil Deo\\
  Johns Hopkins University \\
  \texttt{adeo1@jhu.edu} \\\And
  Kate Sanders\\
  Johns Hopkins University \\
  \texttt{ksande25@jhu.edu} \\\And
  Benjamin Van Durme\\
  Johns Hopkins University \\
  \texttt{vandurme@jhu.edu} \\}
    
\begin{document}
\maketitle
\begin{abstract}
Making theory-of-mind inferences from human dialogue is a strong indicator of a model's underlying social abilities, which are fundamental for adept AI assistants. However, large language and reasoning models struggle to understand sophisticated social phenomena in transcript data, such as sarcasm and irony. To assess the weaknesses of current models and to identify their solutions, we introduce \textsc{SocialNLI} (\textsc{SoNLI})—the first social dialogue \textit{inference} dataset~\footnote{\href{https://huggingface.co/datasets/adeo1/SocialNLI}{SocialNLI dataset.}}.
\textsc{SoNLI} consists of a collection of dialogue transcripts hand-picked to center complex social nuances like irony and sarcasm, paired with inferences, corresponding likelihood scores, and human-written explanations. We explore social inference analysis as a facet of theory-of-mind, and evaluate LLM and reasoning model theory-of-mind ability through multi-step counterfactual reasoning.
\end{abstract}


\section{Introduction}

The ability to observe social circumstances and interpret them in ways that align with human perspectives is a critical skill for large language models (LLMs) and large reasoning models (LRMs). Models must navigate complex social reasoning to function effectively in many environments involving humans, but despite their aptitude for understanding user behavior, contemporary LLMs still falter in understanding and reasoning over various other social scenarios \cite{sarıtaş2025systematicreviewevaluationlarge}. This challenge is rooted in current language models' lack of general theory-of-mind (ToM) capabilities: while models can imitate social understanding through pattern recognition, they often fail to infer the underlying mental states, such as beliefs, emotions, or intentions, that drive human social interactions, especially in cases involving multiple speakers \cite{Amirizaniani_Martin_Sivachenko_Mashhadi_Shah_2024, Sap_LeBras_Fried_Choi_2023, Moghaddam_Honey_2023, Ullman_2023}. ToM stress-tests corroborate this underperformance, as frontier models score lower than humans accuracy on ToM benchmarks and are brittle to minor rephrasings \cite{chen2024tombenchbenchmarkingtheorymind, Shapira_Levy_Alavi_Zhou_Choi_Goldberg_Sap_Shwartz_2023}.

Inference generation from multi-speaker dialogue is a key skill that both demonstrates a model's capacity for ToM reasoning \citep{street2024llmtheorymindalignment, Li_2023}, and can facilitate sophisticated reasoning for downstream tasks~\citep{sanders2025bonsai}. While various benchmarks require abilities surrounding social understanding and inference, there are no text-based datasets targeting the explicit skill of making ToM inferences from dialogue content, and so it is difficult to assess the social reasoning ability of current models and what specific pitfalls they suffer from in their interactions.

\begin{figure}[t]
  \centering
    \includegraphics[width=\columnwidth]{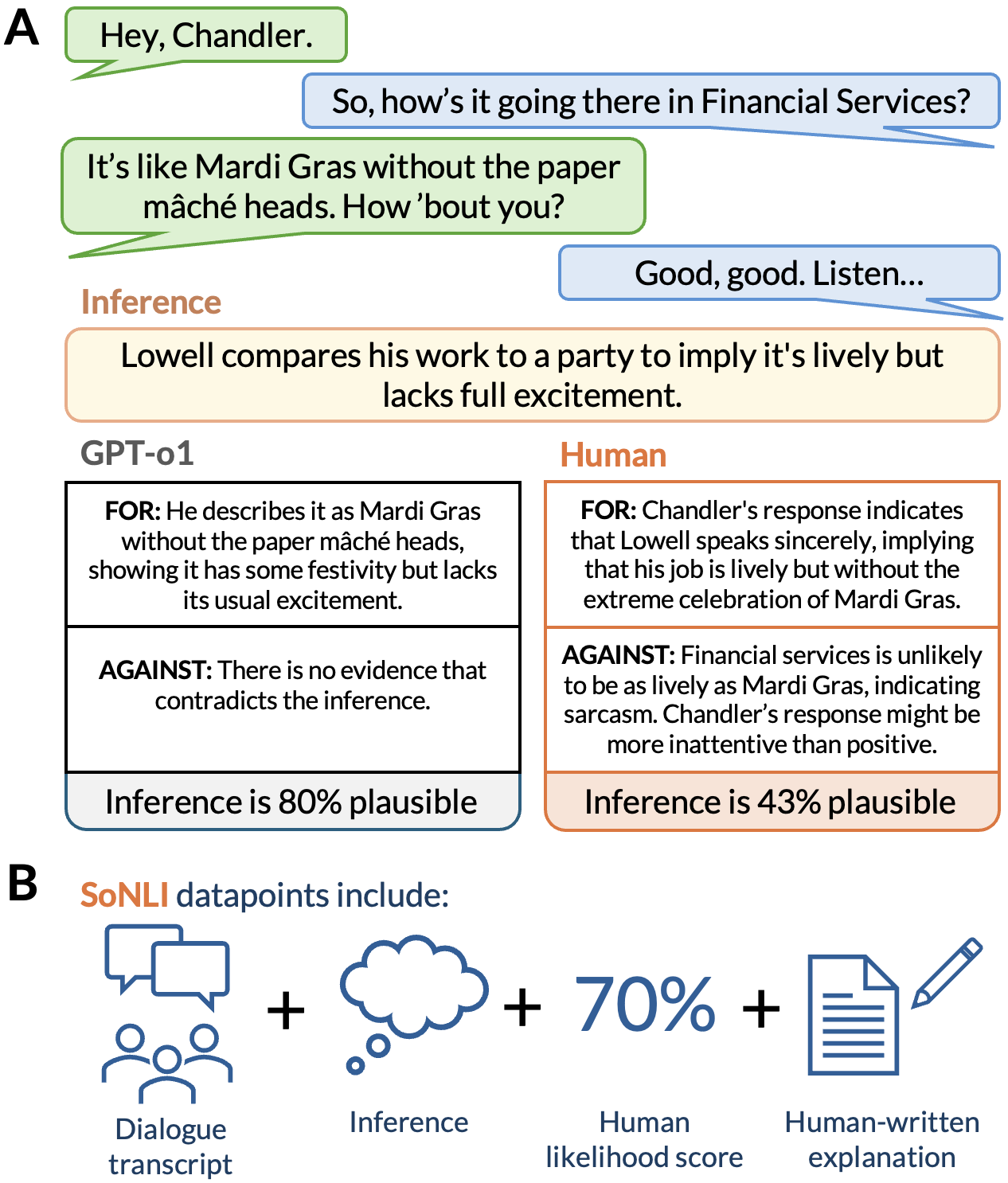}
  \caption{(A) LLMs fall behind humans on counterfactual reasoning over complex dialogue snippets. (B) \textsc{SocialNLI} dataset contents.}
  \label{fig:socialNLI-new-example}
   \vspace{-5mm}
\end{figure}

To address these shortcomings, we present \textsc{SocialNLI} (\textsc{SoNLI}), the first transcript-based social \textit{inference} dataset. Unlike existing NLI resources that focus on factual or multi-domain entailment, \textsc{SoNLI} consists of free-form natural language inferences over dialogue snippets that are hand-selected to target sarcasm and irony. See Fig.~\ref{fig:socialNLI-new-example} for an abridged dialogue sample from \textsc{SoNLI}. Inferences are paired with human-labeled plausibility scores based on the dialogue content as well as human-written natural language explanations for each score. \textsc{SoNLI} then serves both as a benchmark for explicit social ToM evaluation as well as a training and inference resource for socially-aligned LLMs. We demonstrate through counterfactual reasoning experiments that existing models have notable room for improvement on aligning with human social inferences, even when using strong ``thinking model" reasoners. We present:

\begin{itemize}[topsep=0pt, parsep=2pt, leftmargin=*]
  \item \textsc{SoNLI}, a social inference and reasoning dataset centered on sarcasm, irony, and humor, containing both human inference judgments and natural language judgment explanations.
  \item A novel LLM-driven counterfactual inference approach that encourages ToM reasoning to analyze and evaluate social phenomena.
  \item Experiments benchmarking current LLMs on \textsc{SoNLI} to ascertain behavioral discrepancies compared to human annotators and each other.
\end{itemize}

\section{Related Work} 
\paragraph{Natural Language Inference Datasets} 
Following years of RTE~\citep{dagan2005pascal}, \citet{bowman2015largeannotatedcorpuslearning} introduced the influential SNLI benchmark. Multiple popular NLI benchmarks were proposed, motivating more research in the field \citep{Williams_Nangia_Bowman_2018, poliak2018collecting, wang2018glue, wang2019superglue}, alongside a wave of domain‑specific inference datasets including SciTail~\citep{Khot_Sabharwal_Clark_2018}, MedNLI~\citep{Romanov_Shivade_2018}, and XNLI~\citep{conneau2018xnlievaluatingcrosslingualsentence}.
To keep models from over‑fitting to train sets, ANLI iteratively collects examples that models fail on, raising the difficulty of each round of training \citep{Nie_Williams_Dinan_Bansal_Weston_Kiela_2020}. Beyond categorical labels, UNLI models uncertainty as a scalar likelihood, offering an alternative to three-way NLI~\citep{chen2020uncertainnaturallanguageinference}.

\paragraph{Social Reasoning Datasets}
A complementary line of work targets social commonsense directly. A prevalent formulation frames sarcasm and irony detection as classification, with corpora like SARC and its modality and language-specific extensions, such as MUStARD (audio‑visual clips), the English–Arabic iSarcasmEval benchmark, the English-Chinese FanChuan benchmark, and more \citep{Khodak_Saunshi_Vodrahalli_2018, Castro_Hazarika_Pérez-Rosas_Zimmermann_Mihalcea_Poria_2019, Abu_Farha_Oprea_Wilson_Magdy_2022, Zheng_Li_Wu_Ziyi_Hongchao_Hu_Xinjun_Wang_Chen_Luan_et_al._2025, Farabi_2024}.  A more nuanced formulation casts social commonsense as a ToM and belief-tracking task: SocialIQa asks multiple‑choice questions about motivations, emotions, and likely future actions, while higher‑order belief‑tracking benchmarks like HI‑TOM and OpenToM probe deeper ToM reasoning \cite{sap-etal-2019-social, He_Wu_Jia_Mihalcea_Chen_Deng_2023, xu2024opentomcomprehensivebenchmarkevaluating}. Multimodal TV and movie benchmarks assess reasoning over social content as well with a focus on vision~\citep{lei2019tvqalocalizedcompositionalvideo, farabi2024survey}. No existing benchmarks ask models to articulate open-ended inferences derived from conversational cues with an emphasis on complex social dynamics like sarcasm and irony. 


\paragraph{Theory of Mind}
LLM theory of mind (ToM) ability poses many implications for downstream research~\citep{street2024llmtheorymindalignment}, particularly multi-agent collaboration~\citep{Li_2023, kostka2025cognitivesynergyllmbasedmultiagent} and human-in-the-loop interactions~\citep{zhang2024mutualtheorymindhumanai}, including in robotics~\citep{10.1145/3610978.3640767}. Some research suggests that LLMs reflect human performance on ToM tasks~\citep{street2024llmsachieveadulthuman}, while others posit that LLMs struggle on open-ended ToM questions, even after incorporating ToM-specific strategies such as prompt tuning techniques~\citep{10.1145/3627673.3679832}. \citet{riemer2025positiontheorymindbenchmarks} emphasize the distinction between literal and functional ToM, arguing that most benchmarks omit the latter. \citet{shinoda2025tomato} address limited personalities in ToM benchmarks by having LLMs ``act" as different characters.

\section{Dataset}

\subsection{Collecting transcript-question pairs}\label{sec:generating-data}
We require a collection of minimally-processed dialogues to build our annotations on. 
We use the FriendsQA dataset as our basis, a collection of questions about dialogue snippets from the TV show Friends \cite{yang-choi-2019-friendsqa}. Friends is a strong source of dialogue exhibiting sophisticated social reasoning, as it frequently involves sarcasm, irony, and other forms of social nuance. We condition our inference generation on these questions to target these topics.

We hand-select a set of 243 FriendsQA dialogue transcripts that contain elements of sarcasm, irony, or both. We use DeepSeek-R1 to filter out the questions that are unlikely to require interpreting sarcasm or irony, ensuring generated inferences target social nuances (Appendix \ref{sec:social-nuance-target-filtering}). We adopt R1 here as it is a strong reasoning model with high scores on standard reasoning challenges \cite{deepseekai2025deepseekr1incentivizingreasoningcapability}. As FriendsQA was built for open-domain QA, most questions are out-of-scope. As such, we use R1 to generate questions that explicitly require recognizing sarcasm or irony in the dialogue, leaving 532 questions (Appendix \ref{sec:social-nuance-target-question-construction}).

\subsection{Collecting inferences}\label{sec:collecting-inferences}
Aligning with existing NLI datasets, we curate a diverse set of inferences intentionally spanning low– to high–quality. To generate inferences, we prompt GPT-4o with the original dialogue snippet and the corresponding question \cite{openai2024gpt4technicalreport}. We use GPT-4o as it reliably follows instructions and produces coherent long-form text, yielding fluent, diverse hypotheses suitable for scoring \citep{openai2024gpt4ocard}. We use several prompts to allow for a diverse set of inferences, and leverage techniques such as chain-of-thought (CoT) to improve inference quality (Appendix \ref{sec:inference-generation-cot}; \citealp{wei2023chainofthoughtpromptingelicitsreasoning}), while also sampling with GPT-3.5 \emph{without} CoT to intentionally lower reasoning quality (Appendix \ref{sec:inference-generation-no-cot}; \citealp{brown2020languagemodelsfewshotlearners}). We collect 10 inferences per $(d, i, y)$ triple, totaling 5,320 inferences across \textsc{SoNLI}, where $d$ denotes the dialogue, $i$ the inference, and $y\in\{0,1\}$ the scalar plausibility score.

\subsection{Statistics}
As summarized in Table~\ref{tab:dataset-stats}, \textsc{SoNLI} features notably longer and more multi-party dialogues than prior resources. This conversational depth and multi-party interaction demands that models continually track and update multiple mental-state representations—a core requirement for robust ToM reasoning. Thus, \textsc{SoNLI} is among the most demanding social inference and ToM benchmarks to date.

\begin{table}[t]
\centering
\small
\setlength{\tabcolsep}{4pt} 
\begin{tabularx}{\columnwidth}{>{\raggedright\arraybackslash}X cc}
\toprule
Dataset & Turns & Speakers \\
\midrule
\textsc{SoNLI} (all)   & 25.07 & 4.78 \\
\textsc{SoNLI} (eval)  & 24.76 & 5.11 \\
MultiWOZ \cite{budzianowski2020multiwozlargescalemultidomain} & 13.46 & 2 \\
Persona-Chat \cite{zhang2018personalizingdialogueagentsi}     & 6--8  & 2 \\
DailyDialog \cite{li2017dailydialogmanuallylabelledmultiturn} & 7.9   & 2 \\
Multi-party meetings (typically)                                    & --    & 2--4 \\
\bottomrule
\end{tabularx}
\caption{Average turns and speakers per dialogue}
\label{tab:dataset-stats}
\vspace{-15pt}
\end{table}

Additionally, across \textsc{SoNLI}, R1 assigns a mean plausibility of 0.823 (median$=$1.0, $\sigma=0.283$)  The 1.4k human-evaluated subset also averages 0.823, but annotators’ scores average $0.863\pm0.172$.

\subsection{Annotating the data}\label{sec:annotating-data}
For the evaluation set, we collect human plausibility scores and natural language explanations for 1,400 inferences over 70 dialogue transcripts using Amazon Mechanical Turk. We first conduct a series of pilot tasks to curate a group of high-quality human annotators, testing both for quality of labels and human-ness. The annotator pool is limited to annotators in the U.S. with over 1000 HITs accepted and an acceptance rate of >98\%.\footnote{We pay annotators an average of \$15/hr.} Annotators were presented with dialogue snippets containing complex social reasoning elements, along with corresponding inferences, and were instructed to rate the plausibility of each inference on a sliding scale from ``0\% true" to ``100\% true" and explain their ratings with up to a few sentences. The full task instructions are included in Appendix~\ref{sec:human-annotations}. Human annotators show a positive bias (mean=0.863, median=0.9, $\sigma$=0.172), with 54.9\% of scores landing above 0.8. This skew is expected because we curate socially plausible inferences, meaning that models must resolve fine-grained differences to perform well on \textsc{SoNLI}.

\section{Experiments}
We target two primary research questions: \textbf{(R1)} How well do model-produced plausibility scores for open-ended social inferences align with human-graded judgements? \textbf{(R2)} To what extent is model performance influenced by their argument generation ability vs. their plausibility scoring ability?

\subsection{Counterfactual reasoning}\label{sec:counterfactual-reasoning}
By design, most inferences in \textsc{SoNLI} are neither objectively true or false. We introduce an UNLI-inspired, counterfactual approach to LLM-driven reasoning over complex, ToM-centric social reasoning questions. For a social inference \(I_i\), we elicit two synthetic counterfactual arguments—\(A_i^+\) arguing why \(I_i\) is true and \(A_i^-\) arguing why it is false. These arguments correspond to the likelihoods
\[
\begin{aligned}
s_i^+ &= P(A_i^+ \mid I_i=\text{true})\\
s_i^- &= P(A_i^- \mid I_i=\text{false})
\end{aligned}
\]
Assuming a uniform prior on \(I_i\) and conditional independence of \(A_i^+\) and \(A_i^-\) given \(I_i\), we compute the posterior probability that \(I_i\) is true via
\[
\hat{P}\bigl(I_i=\text{true}\mid A_i^+,A_i^-\bigr)
= \tfrac{\hat{s_i}^+(1 - \hat{s_i}^-)}{%
     \hat{s_i}^+(1 - \hat{s_i}^-)+(1 - \hat{s_i}^+)\hat{s_i}^-}.
\]\label{eq:post}
We release DeepSeek-R1-generated counterfactual arguments and scores for each inference as a part of \textsc{SoNLI} to aid future training and analysis.
\subsection{Model–Human Alignment}\label{sec:exp1}

\paragraph{Setup} In this experiment we target \textbf{R1}: How LLM plausibility scores on \textsc{SoNLI} align with humans. We evaluate on the \textsc{SoNLI}'s 1.4k human-scored evaluation set of 1.4k inferences. For each model under test (GPT-4o, DeepSeek-V3, Llama-3.1-70B-Instruct, Qwen2.5-32B, Llama-3.1-8B-Instruct, GPT-4o-mini), the model \emph{itself} generates the supporting and opposing counterfactual arguments, assigns their heuristic likelihoods, and then computes the final plausibility $\hat p_{\mathrm{CR}}$ using Eq.~\eqref{eq:post}. We compare $\hat p_{\mathrm{CR}}$ to human scores $y$ using Pearson’s correlation $\rho$ and mean absolute error (MAE). Prompts are in the appendix.

\vspace{-1pt}
\paragraph{Results} 
\begin{table}[t]
\small
  \centering
  \setlength{\tabcolsep}{4pt}
  \resizebox{\columnwidth}{!}{%
      \begin{tabular}{lccc}
        \toprule
        \textbf{Model} & $\boldsymbol{\rho}$ & \textbf{MAE} \\
        \midrule
        DeepSeek-V3               & -0.137 & 0.294 \\
        GPT-4o                    & \textbf{0.153} & 0.341 \\
        GPT-4o-mini               & 0.095 & 0.186 \\
        Llama-3.1-8B-Instruct     & 0.010 & 0.359 \\
        Llama-3.1-70B-Instruct    & 0.075 & 0.185 \\
        Qwen2.5-32B-Chat            & -0.028 & \textbf{0.181}  \\
        \bottomrule
      \end{tabular}
  }
  \caption{\textbf{End-to-end counterfactual reasoning plausibility alignment on \textsc{SoNLI}.} For each entry, the evaluated model generates supporting and opposing counterfactual justifications, generates likelihoods for each, and produces a plausibility score using the method in Section ~\ref{eq:post}. We report Pearson correlation $\rho$ with human scores, and mean absolute error (MAE).}
  \label{tab:exp1_results}
  \vspace{-15pt}
\end{table}

Table~\ref{tab:exp1_results} shows that only GPT-4o, and to some extent GPT-4o-mini, exhibit a meaningful positive correlation with human scores, indicating some ability to separate plausible from implausible inferences. Most other models sit near zero or negative $\rho$, suggesting weak or inverted agreement with annotators. Several models nonetheless post low MAEs by placing scores in a tight band near the dataset mean, most notably Qwen2.5-32B and Llama-3.1-70B, reducing magnitude error without preserving instance-wise ordering. Accordingly, we treat $\rho$ as the primary evaluation signal for \textsc{SoNLI} and use MAE as a complementary calibration indicator.

\subsection{Human-Judged Counterfactuals}\label{sec:exp2}
\paragraph{Setup} In this experiment, we rate arguments with human judges, and
address \textbf{R2} by ablating the quality of arguments generated by LLMs and LRMs 
on \textsc{SoNLI}. We have humans fact-check model-generated supporting and opposing explanations for each inference (Section~\ref{sec:counterfactual-reasoning}) to see whether the low performance in Section 4.2 was due to poor argument scoring or poor argument generation. We test three LLM-LRM pairs: GPT-4o/o1, Qwen2.5-32B/QwQ-32B, and DeepSeek-V3/DS-R1, on 27 inferences drawn from 9 dialogues (Section~\ref{sec:annotating-data}; \citealp{openai2024openaio1card, qwq32b, deepseekai2025deepseekr1incentivizingreasoningcapability}). All inferences and dialogues are verified to have sarcasm or irony.

\begin{figure}[t]
\centering
\begin{tikzpicture}
\begin{axis}[
    width=\columnwidth,
    height=0.7\columnwidth,
    ybar,
    bar width=20pt,
    ylabel={Overall Accuracy (\%)},
    ylabel style={yshift=-8pt},
    ymin=60,
    ymax=100,
    symbolic x coords={Human, GPT, Qwen, DeepSeek},
    xtick=data,
    xticklabels={GPT-4o / \\ o1, Qwen2.5-32B / \\ QwQ-32B, DS-v3 / \\ DS-R1},
    xticklabel style={
        font=\scriptsize,
        align=center
    },
    enlarge x limits=0.22,
    nodes near coords,
    nodes near coords style={font=\scriptsize, yshift=-2.5pt},
    grid=major,
    major grid style={dashed,gray!30},
    legend style={
        at={(0.02,0.98)},
        anchor=north west,
        legend columns=1,
        font=\scriptsize
    },
]

\addplot[fill=blue!60] coordinates {
    (GPT,80.39)
    (Qwen,73.58)
    (DeepSeek,79.63)
};

\addplot[fill=orange!60] coordinates {
    (GPT,85.19)
    (Qwen,78.26)
    (DeepSeek,75.93)
};

\draw[red, thick] (rel axis cs:0,0.82375) -- (rel axis cs:1,0.82375); \node[anchor=west, font=\scriptsize] at (rel axis cs:1,0.82375) [xshift=5pt] {Human (92.59\%)};
\end{axis}
\end{tikzpicture}
\vspace{-3mm}
\caption{\textbf{Human-judged factuality of explanations.} For each inference, models produce supporting and opposing explanations, and humans score the accuracy of those explanations. LLMs are blue; LRMs are orange. The red horizontal line shows \textbf{human baseline performance (92.59\%)}, exceeding all model accuracies.}
\label{fig:overall-correctness}
\vspace{-12pt}
\end{figure}
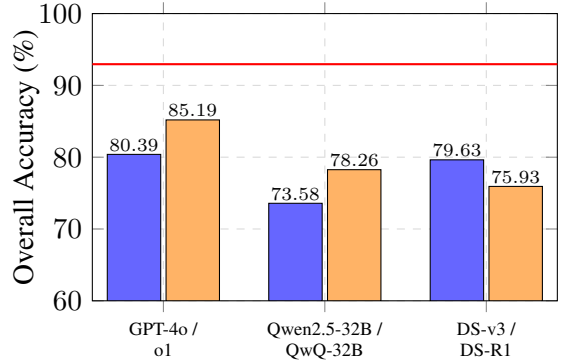

\paragraph{Results}
LRMs are rated higher than LLMs: average factuality 79.8\% vs.\ 77.9\%, with o1 at 85.2\% and GPT-4o at 80.4\% (see Fig.~\ref{fig:overall-correctness}). Despite the edge, all evaluated models under-perform humans by 7.40\%-19.01\%. With Experiment~1’s weak model–human correlations, these results show that current systems remain far from robust ToM reasoning, although models may struggle more with weighing arguments than generating them.

\section{Conclusion}
We introduce a natural language inference dataset for interpreting high level social phenomena like sarcasm and irony from dialogue transcripts, including both detailed human annotations and various LLM and reasoning model outputs. Through evaluations, we show that current language and reasoning models struggle to demonstrate social awareness on the dataset inference task, showing there is significant room for improvement on developing models capable of ToM reasoning and deep social understanding.

\section*{Limitations}
Our dataset focuses on dialogue transcripts from one TV show. While Friends covers a wide range of sophisticated social reasoning topics, it is not representative of social phenomena as a whole. Related, our dataset is only in English and predominantly tackles social reasoning from an English-speaking perspective. This should be taken into account, to be aware of potential biases propagated by this dataset.

Our experiments cover a relatively small set of data points due to the human labor involved in evaluations. We view this dataset as a diagnostic tool to prompt further investigations into model behavior in larger-scale settings.

Finally, while we focus on human evaluations as models are not sufficiently adept to judge outputs themselves, the human annotators inherently have social biases themselves. The redundancy of these annotations is low to account for cost, but we provide all annotation instructions to allow for more labels to be collected to ensure more redundancy in future work.

In this paper, all artifacts are used appropriately for research purposes.

\bibliography{custom}

\appendix
\section{Dataset Card}
\subsection{Overview}
\textsc{SocialNLI} (\textsc{SoNLI}) is a dialogue–centric benchmark for open-ended social inference. Each example pairs a short multi-speaker transcript with a free-form hypothesis and asks models to assign a scalar plausibility score in $[0,1]$, optionally accompanied by brief supporting and opposing rationales. Dialogues are drawn from FriendsQA and were curated to foreground sarcasm, irony, and related social cues. The release includes two splits: an \emph{auto/main} split of 3{,}920 LLM-generated and scored inferences, and an \emph{eval} split of 1{,}400 human-scored inferences over 70 dialogues (5{,}320 total). Each record contains the dialogue text, the hypothesis, model and/or human scores, and counterfactual explanations; the eval split additionally provides annotator justifications. The dataset is intended for evaluating theory-of-mind–style social reasoning and explanation quality.

\subsection{Inference Classifications}

In addition to the dataset processing and generation conducted in Sections~\ref{sec:generating-data} and \ref{sec:collecting-inferences}, we also attempt to see what types of inferences are generated. Leveraging the prompt described in Section~\ref{sec:inference-classification}, we classify each inference in \textsc{SoNLI} as either concerning reality, a belief, or an emotion. We use these coarse labels chiefly as an auxiliary data-point to monitor diversity of inferences type.

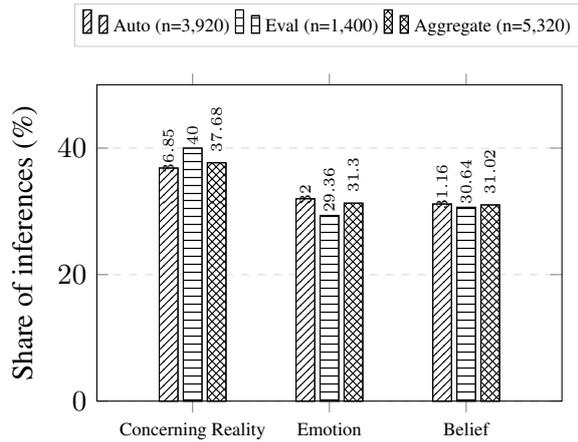
\begin{figure}[t]
  \centering
  \begin{tikzpicture}
    \begin{axis}[
      width=\columnwidth,
      height=0.75\columnwidth,
      ybar,
      bar width=7pt,
      ymin=0, ymax=50,
      ylabel={Share of inferences (\%)},
      symbolic x coords={Concerning Reality, Emotion, Belief},
      xtick=data,
      xticklabel style={font=\scriptsize},
      enlarge x limits=0.35,
      ymajorgrids,
      major grid style={dashed,gray!30},
      clip=false,
      legend style={
        font=\scriptsize,
        at={(0.5,1.12)}, anchor=south,
        legend columns=3,
        fill=white, draw=black!20
      },
      nodes near coords,
      nodes near coords style={font=\tiny, inner sep=1pt, anchor=west, rotate=90},
    ]

      \addplot[
        draw=black, fill=white,
        postaction={pattern=north east lines, pattern color=black},
        every node near coord/.append style={xshift=-3pt}
      ] coordinates {
        (Concerning Reality,36.85)
        (Emotion,32)
        (Belief,31.16)
      };
      \addlegendentry{Auto (n=3{,}920)}
      
      \addplot[
        draw=black, fill=white,
        postaction={pattern=horizontal lines, pattern color=black}
      ] coordinates {
        (Concerning Reality,40)
        (Emotion,29.36)
        (Belief,30.64)
      };
      \addlegendentry{Eval (n=1{,}400)}
      
      \addplot[
        draw=black, fill=white,
        postaction={pattern=crosshatch, pattern color=black},
        every node near coord/.append style={xshift=3pt}
      ] coordinates {
        (Concerning Reality,37.68)
        (Emotion,31.3)
        (Belief,31.02)
      };
      \addlegendentry{Aggregate (n=5{,}320)}

    \end{axis}
  \end{tikzpicture}
  \caption{\textbf{Inference type distribution}.}
  \label{fig:inference-type-dist}
\end{figure}

Figure~\ref{fig:inference-type-dist} shows the distribution of inference types across all splits. For the auto split (3,920 items), inferences are classified as concerning reality (36.85\%), emotion (32.00\%), and belief (31.16\%). The eval split (1,400 items) shows a similar distribution: concerning reality (40.00\%), emotion (29.36\%), and belief (30.64\%). Aggregating across all 5,320 labeled items yields concerning reality (37.68\%), emotion (31.30\%), and belief (31.02\%). These results demonstrate that all three inference categories are roughly balanced across splits, with inferences concerning reality slightly more prevalent in each case.

\section{Human Annotations}\label{sec:human-annotations}
No annotator information was recorded in the annotation process.
\subsection{Main instructions}
\begin{instrbox}\small
In this task, you are presented with a short dialogue transcript taken from a TV show episode.
Please read all of the transcript. Don't worry if some of the text is transcribed poorly, your best interpretation is fine.
Then, for each statement below about the transcript, mark whether it is most likely true or false using the slider.
The slider is on a 100-point scale, meaning that "0" is absolutely false and "100" is absolutely true.
"50" indicates that there is equal probability that the statement is true or false.
After selecting the most accurate likelihood score for the statement based on the transcript, write a brief explanation for your score (about one sentence).
There will be twenty statements per dialogue transcript.\\

When scoring an inference, it’s important ask yourself a few questions:

\begin{itemize}
\item Is the inference true?
\item Is the inference relevant to the dialogue transcript?
\item Is the inference vague or non-descriptive?
\item Does the inference introduce information that isn’t included in the dialogue transcript?
\end{itemize}

Please check the instructions for a quick example. Thanks!\\

Thank you for participating in the task. Quality of submitted tasks will be carefully monitored, and bot outputs will be rejected.
\end{instrbox}

\subsection{Score Processing and Precision}

We collect human plausibility scores on a 0--100 slider and then normalize them to $[0, 1]$. To avoid false precision from the slider UI, and human annotators in general, we quantize normalized scores to one decimal place in our statistical analyses. To balance fidelity, \textsc{SoNLI} retains scores of up to two significant figures, giving downstream users flexibility to choose their preferred granularity.

\subsection{Example}
\subsubsection{Dialogue transcript}
\begin{transcriptbox}\small
\speaker{Phoebe Buffay} Wow ! I can not believe Mark asked you out .\\
\speaker{Rachel Green} I know .\\
\speaker{Phoebe Buffay} What , so what are you gonna tell him ?\\
\speaker{Rachel Green} Well , I told him I would think about it , but I 'm gonna tell him no .\\
\speaker{Rachel Green} I mean I think I 'd say no to anybody right now . Oh , but it was so strange . I mean I 'm standing there with this charming , cute guy , who 's asking me to go out with him , which I 'm allowed to do , and I felt guilty . Y'know , like I 'd be cheating on Ross or something .\\
\speaker{Phoebe Buffay} Wow . So , okay , maybe that means that , you 're not over Ross yet and you have issues with your father .\\
\speaker{Rachel Green} I do n't have any issues with my Father .\\
\speaker{Phoebe Buffay} Okay , so it 's probably just the Ross thing then .
\end{transcriptbox}
\subsubsection{Example inferences}

\begin{infcard}
\textbf{Inference:} Rachel feels guilty about dating Mark because she still has feelings for Ross.

\begin{infcheck}
\item True? \quad Yes – Rachel explicitly says she felt guilty “like I’d be cheating on Ross.”
\item Relevant? \quad Yes – it directly explains her emotional reaction in the transcript.
\item Non-vague? \quad Yes – pinpoints what she feels (guilt) and why (Ross).
\item Extra info? \quad No – everything is stated or implied in her lines
\item Verdict: Good inference – score high
\end{infcheck}
\end{infcard}

\begin{infcard}
\textbf{Inference:} Rachel has commitment issues with all men.

\begin{infcheck}
\item True? \quad Not sure.
\item Relevant? \quad No – the provided dialogue only covers this one date invite, not “all men.”
\item Non-vague? \quad No – the inference isn’t very specific
\item Extra info? \quad Yes – nothing in the dialogue indicates “commitment issues.”
\item Verdict: Bad inference – score low
\end{infcheck}
\end{infcard}

\section{Prompts}
\subsection{Generation Parameters}\label{sec:generation-parameters}
All prompts were executed with temperature equaling 0.7, and maximum tokens equaling 5000 (4096 for GPT-3.5). These settings were applied consistently across all inference generation, explanation generation, and classification tasks to ensure reproducibility.
\subsection{Social Nuance Target Filtering}\label{sec:social-nuance-target-filtering}
\begin{promptbox}
<role> You are an AI assistant specializing in analyzing the nuances of language in questions and dialogues. </role>

<task> Your task is to analyze the provided question, and the accompanying dialogue, to determine the likelihood that understanding sarcasm or irony in the dialogue is necessary to answer the question accurately. If the question doesn't require understanding sarcasm or irony, choose "very unlikely" or "unlikely", and justify your choice. If the question requires understanding sarcasm or irony, choose "very likely" or "likely", and justify your choice. If you are unsure or it is possible, choose "possibly" and justify your choice.</task>

<response_format> You MUST choose your answer exclusively from the following five categories:
- very unlikely
- unlikely
- possibly
- likely
- very likely

Wrap your thoughts in <think> </think> tags and your final answer in <answer> </answer> tags.
</response_format>

<dialogue>
{dialogue}
</dialogue>

<question> {question} </question>
\end{promptbox}

\subsection{Social Nuance Target Question Construction}\label{sec:social-nuance-target-question-construction}
\begin{promptbox}
<role> You are an AI assistant specialized in understanding and identifying sarcasm and irony in conversations. Your goal is to create insightful questions that probe the sarcastic or ironic elements within a dialogue. </role>

<task> Your task is to carefully read the provided dialogue and generate one high-quality question that specifically targets or requires understanding of the sarcasm or irony present in the dialogue. The question should make sense in the context of the dialogue and should not be answerable without recognizing the sarcastic or ironic intent.

Ensure the question is:
- Grammatically correct and clearly phrased.
- Directly related to the sarcastic or ironic content.
- Open-ended or requires more than a simple yes/no answer if possible, encouraging deeper reasoning about the sarcastic/irony.
- Not a repeat of existing questions (if any were implicitly provided or are obvious).
</task>

<dialogue>
{dialogue}
</dialogue>

<response_format>
Provide only the generated question. Do not include any preamble, explanation, or XML tags in your response.
For example, if the dialogue implies someone is "thrilled" about bad news sarcastically, a good question might be: "What does Speaker A *really* think about the news?" or "How does Speaker B's tone when saying 'fantastic' suggest their true feelings?"
</response_format>

<question_output_placeholder>
[Your generated question here]
</question_output_placeholder>
\end{promptbox}

\subsection{Inference Generation (CoT)}\label{sec:inference-generation-cot}
\begin{promptbox}
**Instructions**: For the following question and scene dialogue, write five distinct and non-overlapping inferences entailed by the scene. Generate the inferences by reasoning through the scene dialogue and the question in a step by step manner. Ensure the inferences are logical and relevant to the scene. The inferences should resemble short, factual statements about the scene and should help answer the question using component reasoning steps.

**Output Format:**
Wrap the reasoning process in <think> </think> tags and the inferences with ```json ``` markdown tags. The inferences should be in JSON format with keys "1" through "5". Do not include any additional text, explanations, or formatting in the JSON.

**Example Output Format:**
<think>
<reasoning step 1>
<reasoning step 2>
...
</think>
```json
{{
  "1": "Inference one.",
  "2": "Inference two.",
  "3": "Inference three.",
  "4": "Inference four.",
  "5": "Inference five."
}}```

**Question**: {question}

**Scene**: {scene_dialogue}

**Inferences (5 total)**:
\end{promptbox}

\subsection{Inference Generation (no CoT)}\label{sec:inference-generation-no-cot}
\begin{promptbox}
**Instructions**: For the following question and scene dialogue, write five distinct and non-overlapping inferences entailed by the scene. The inferences should resemble short, factual statements about the scene and should help answer the question using component reasoning steps.

**Output Format:**
Return the inferences in JSON format with keys "1" through "5". Do not include any additional text, explanations, or formatting.

**Example Format:**
```json
{{
  "1": "Inference one.",
  "2": "Inference two.",
  "3": "Inference three.",
  "4": "Inference four.",
  "5": "Inference five."
}}```

**Question**: {question}

**Scene**: {scene_dialogue}

**Inferences (5 total)**:
\end{promptbox}

\subsection{Supporting Explanation Generation}\label{sec:supporting-explanation-generation}
\begin{promptbox}
<role> You are an AI assistant specializing in creating concise, evidence-based explanations to support inferences. </role>

<task> Create an explanation that directly supports the inference. Find and cite specific evidence that directly supports the inference. Focus on relevant details that prove the inference is true. If there is no evidence that supports the inference, simply state that there is no evidence that supports the inference. Do not repeat the inference or dialogue in the explanation. </task>

<tone> The explanation should be simple, concise and declarative. It should be a single sentence that directly supports the inference. </tone>

<format> Write your thoughts in <think> </think> tags. Do not include any additional text, explanations, or formatting in the answer. </format>

<dialogue>
{scene_dialogue}
</dialogue>

<inference> {inference} </inference>

<think>
\end{promptbox}

\subsection{Opposing Explanation Generation}\label{sec:opposing-explanation-generation}
\begin{promptbox}
<role> You are an AI assistant specializing in creating concise, evidence-based explanations to prove inferences false. </role>

<task> Create an explanation that proves that the inference is false. Find and cite specific evidence that directly contradicts the inference. Focus on relevant details that prove the inference is false. If there is no evidence that contradicts the inference, simply state that there is no evidence that contradicts the inference. Do not repeat the inference or dialogue in the explanation. </task>

<tone> The explanation should be simple, concise and declarative. It should be a single sentence that opposes the inference. </tone>

<format> Write your thoughts in <think> </think> tags. Do not include any additional text, explanations, or formatting in the answer. </format>

<dialogue>
{scene_dialogue}
</dialogue>

<inference> {inference} </inference>

<think>
\end{promptbox}

\subsection{LLM-as-a-Judge for Explanations}\label{sec:llm-as-a-judge}
Prompt adapted from~\citet{sanders2025bonsai}.
\begin{promptbox}
You are a reasoning system that analyzes the likelihood of complex events given information about hypothetical scenarios. 

You are given a description of a fictional scenario and a hypothesis about that scenario that may or may not be true. Given the situation, you will first score the likelihood that this hypothesis is true, on a scale from 0 to 10, using the following rubric as guidance:

0 (virtually impossible): Essentially no way the hypothesis could possibly be true, given the evidence. Less likely than being struck by lightning.
1 (unlikely): The hypothesis is unlikely, but definitely not impossible.
2 (possible): The hypothesis could be true given the evidence, but there is better chance that it is false. Less likely than drawing a card of the suit of clubs from a standard card deck.
3 (reasonable chance): You would not be more than mildly surprised that the hypothesis is true. About one thirds chance.
4 (a bit less than even-odds): Slightly below fifty-fifty probability. You would not bet more than a small sum that the hypothesis is false.
5 (fifty-fifty): Given the information about the situation, there is approximately equal chance that the hypothesis is true vs. the hypothesis is false. As likely as a fair coin landing on heads.
6 (a bit more than even-odds): Slightly above fifty-fifty probability. You would not bet more than a small sum that the hypothesis is true.
7 (probable): Likely, but you would still not be overly surprised if the hypothesis turned out to be false.
8 (quite likely): About as likely as \*not\* rolling a ``2" with a six-sided die.
9 (extremely likely): Quite certain. You would bet a large amount of money on the hypothesis being true.
10 (practically certain): You cannot imagine a scenario in which the hypothesis is not true, given the situational evidence.

Label your initial prediction with (0), and label your updated predictions with the evidence number it corresponds to. Write your enumerated explanations and probability scores, and nothing else.

Here is a first example:

ORIGINAL DESCRIPTION: There were puddles in the street and dark clouds hung overhead. The Mississippi flag was visible on a nearby car.

HYPOTHESIS: A tornado rolled through a town in Mississippi.

EXPLANATION: It is more likely this was just a regular rainstorm than a tornado. While it might be in Mississippi, there's not enough evidence to claim a tornado occurred.

SCORE: 1

Here is a second example:

ORIGINAL DESCRIPTION: There is a large crowd of people gathered before a lit-up stage at night.

HYPOTHESIS: The band Blur performed at Coachella 2024.

EXPLANATION: While a big nighttime show with a large crowd could describe many concerts or festivals, there is no direct evidence that this specific event is Coachella or that the band on stage is Blur.

SCORE: 2

That is the end of the examples. Now, it's time for you to assign probabilities to a new fictional scenario:

ORIGINAL DESCRIPTION: {explanation}

HYPOTHESIS: {inference}

EXPLANATION:
\end{promptbox}

\subsection{Inference Classification}\label{sec:inference-classification}
\begin{promptbox}
**Instructions**: For the following question and scene dialogue, classify the provided inference as either concerning reality, a belief, or an emotion. Reason through the inference, the question and the scene dialogue step by step, and determine the most appropriate classification. Provide a brief explanation for your reasoning.

**Output Format:**
Wrap the reasoning process in <think> </think> tags and the classification in <answer> </answer> tags. The answer should be one of the three values: "concerning reality," "belief," or "emotion." Do not include any additional text, explanations, or formatting in the answer.

**Example Output Format:**
<think>
<reasoning step 1>
<reasoning step 2>
...
</think>
<answer>
concerning reality
</answer>

**Question**: {question}

**Scene**: {scene_dialogue}

**Inference**: {inference}

**Inference Classification and Reasoning**:
\end{promptbox}

\section{Licensing}
The FriendsQA dataset is available under the Apache License, Version 2.0.\footnote{https://github.com/emorynlp/FriendsQA/tree/master}

\end{document}